# Predicting Startup Exit from Textual Descriptors: A Computational Linguistics Framework

Alberto M.G. Saruggia
Independent Researcher
MBA Candidate, Boston University

Sébastien Germano
Independent Researcher
MSc Candidate in Data & AI, ECE Paris

***Abstract*** – This study shows that textual descriptors alone can predict early-stage startup success, defined as Exit, without relying on contextual, financial, or human capital variables. Using venture capital-curated datasets covering 7,419 startups over 20 years, the research isolates text-based framing variables and engineers 850 features via startup narrative mapping. Data subsets and vector embeddings are evaluated for statistical significance, followed by supervised machine learning experiments across six models. Binary Exit prediction using Logistic Regression attains an F1 of 0.48 with 0.55 recall using all features (excluding embeddings), and an F1 of 0.26 with 0.59 recall using textual descriptors only (including embeddings). Feature analysis indicates that optimized densities of hyping markers such as adjectives, jargon, and buzzwords are associated with higher Exit probability, while excessive statement or name length is associated with lower probability. The study also introduces a quantifiable Hyping Score for potential application in venture screening. Findings indicate that startup framing can serve as standalone predictor of economic outcomes, in high-information-asymmetry investment environments.

***Keywords*** *– computational linguistics, machine learning, information asymmetry, startup success, venture capital*

## I. INTRODUCTION

The language used by early-stage entrepreneurs looking for venture capital (VC) is often inflated and not convincing. This is certainly problematic in a marketplace where investors operate under high information asymmetry, relying on founder narrative, and where only a few relevant parties are available for each side. With such limitations, the matching process is heavily influenced by aligned perceptions of risk (POLZIN 2018). **Investment pitch** is core phase of the matching process where founders without observable track-record have the difficult task of picturing both disruptive potential and a corresponding business model.

A key pitch tactic is the startup **framing**, a linguistic device aimed at highlighting selected aspects of the startup's perceived reality (LO 2022). Since narrative influences resource acquisition above and beyond factual information (MARTENS 2007), founders are incentivized to manipulate the framing in order to translate VC expectations into legitimacy and access to resources (HAMPEL 2025). The degree of such manipulation is determined by risk tolerance and entrepreneurial archetypes (HUBER 2021), whereas confident founders frame their venture's future in abstract and aspirational ways, pitching it as "Amazon-of-X" or "Uber-but-for-Y". They are illustrating their ambition rather than concrete capabilities early on, and in doing so, they are pitching not only their ventures but also themselves. Over the years this approach to framing has been reinforced by mantras like "fake it until you make it" and "fail fast", implicitly promoting narrative amplification as a proven success tactic. Entrepreneurs interpret these slogans as a license for poor experiments and uncalculated risks, eventually leading to hyping and fabrications (GARUD 2014).

While framing is a deliberate tactic, **signals** are factual and costly anchor points that function as proxies for legitimacy or market readiness (CONNELLY 2011). Signals can be directly observed throughout a pitch deck narrative, such as in the financials metrics (e.g., revenues, funding rounds, assets owned), human capital (e.g., team size and education, patents) and business context (e.g., founding year, headquarter location, industry segments).

In this context, **hype** is "the collective vision and promise of a possible future, around which attention, excitement and expectations increase over" (LOGUES 2022), or, in other words, the assumption that a venture may achieve exorbitant growth in a short period of time (LAZAROW 2020). Hyping occurs when founders intensively inflate signals with claims of superiority or uniqueness – "the next big thing" – in an attempt to generate, or be part of, a hype cycle (FENN 2008). Hyping has multiple negative effects, including role-playing (HARTMANN 2022), and deceit framing (GARUD 2025), but it represents also an opportunity, as new ventures use it as a cultural resource to align their goals with the expectations of the hype. Notably, the hype curve only represents 1 in 5 innovations, suggesting that most VC firms and founders consider hype as a potential risk (ROYAL-LAWSON 2024). Between 75% and 94% of startups fail (POLLMAN 2023), underscoring how important winners are to VC and how difficult it is to predict startup success, due to its hyped nature.

**This study shows that startup framing possesses standalone predictive power for startup success, even without considering other variables**. The analysis quantifies: (1) probability thresholds at which text-based hyping correlates with Exit; and (2) predictive power of startup descriptors (name, statement and website URL), both independently and when combined with context variables.

## II. RESEARCH CONTRIBUTION

A substantial body of research has addressed the challenge of predicting startup success. A cohort of ten of such studies is presented to evaluate their assumptions and potential gaps.

TABLE I.

| Comparison of selected literature in the context of signaling and framing | | | | | | | |
|---|---|---|---|---|---|---|---|
| | | Signaling (factual anchor points) | | | Framing (linguistic devices) | | Target Variable |
| ***Research*** | ***Data Source*** | ***Context*** | ***Financials*** | ***Human Capital*** | ***Social & Media*** | ***Descriptors*** | ***Startup Success*** |
| Xiang 2012 | Crunchbase | x | x | x | x | | M&A |
| Guzman 2015 | Government | x | | x | | x | M&A, IPO |
| Sharchilev 2018 | Crunchbase | x | x | x | x | x | Funding rounds |
| Cemre 2019 | Crunchbase | x | x | | x | | M&A, IPO, Operating |
| Kaiser 2020 | Government | x | x | x | | x | M&A, IPO, Operating |
| Garkavenko 2022 | Government | x | x | x | x | | Funding rounds |
| Gavrilenko 2022 | Crunchbase | x | x | | x | | Funding rounds |
| Maarouf 2023 | Crunchbase | x | x | x | | x | M&A, IPO, Funding Rounds |
| Gavrilenko 2024 | Crunchbase | x | x | x | x | x | Funding rounds |
| Bidgoli 2024 | Crunchbase | x | x | x | x | | Funding rounds |

A summary of this review is presented in TABLE I. All studies are structured around public data sources, acquired either from government agencies or the startup directory Crunchbase. The definition of startup success presents main differences: most papers intend success broadly, including funding rounds and continued operations (SHARCHILEV 2018; GARKAVENKO 2022; GAVRILENKO 2022), while only two intend it strictly to the company being acquired or public (XIANG 2012; GUZMAN 2015). The independent variables are categorized into signaling variables (Context, Financials, and Human Capital) and framing variables (Social & Media and Descriptors). Among the studies considering framing variables, two focus on Descriptors (KAISER 2020; MAAROUF 2024), while three on Social & Media (CEMRE 2019; GAVRILENKO 2023; BIDGOLI 2024). No research has been found considering framing variables only. This research builds on this literature – and in particular on the concept of framing (LO 2022) – to address gaps in data sourcing, target variable, and independent variables. First, data are sourced from VC firms rather than public directories, ensuring that inputs by founders are externally validated. Second, startup success is defined only as a cash-returning event, excluding proxies such as funding rounds that may dilute results. Third, textual descriptors are isolated to evaluate their standalone predictiveness. The goal is to analyze a real-world scenario where VC firms align startup narratives to success rationale.

The rest of this research is organized as follows. Section III presents an overview of the data and their key attributes. Section IV investigates experiments and analysis to demonstrate the subject. Section V explores potential business applications. Section VI concludes.

## III. DATA

### A. *Acquisition and preparation*

Two publicly available CSV files downloaded from Kaggle contain scraped data from Techstars and Y Combinator, major VC firms based in U.S. with similar profiles: both founded in 2005-2006, investing in early-stage startups, and with a track-record of notable Exits. The data align with our objectives as follows:

- **Breadth:** all required variables are present.
- **Depth:** 7,419 startups allow a diverse dataset.
- **Completeness:** 93.6% of non-blank cells.
- **Representativeness:** multiple startup cohorts are represented (20 years, 93 countries, 60 industries).

### B. *Variable Analysis*

After pre-processing, the input fields are classified into five variable categories: Descriptors, Context, Social & Media, Human Capital and Exit. Only Descriptors and Context are retained for startup narrative mapping shown in TABLE II, used to align these variables to signaling and framing respectively. **Context Variables** are discrete fields based on a set of predefined categories. They signal factual anchor points and attributes that are mutually exclusive. **Descriptors Variables** are unstructured strings of natural language text and represent the linguistic devices used by founders to frame their project.

TABLE II.

| Startup Narrative Mapping | |
|---|---|
| Signaling Variables | Framing Variables |
| ***Context (category)*** | ***Descriptors (text)*** |
| Founding year | Name |
| Country | Statement |
| Industry Tag | Website URL |

In this research, **Exit** is a binary target variable and means only M&A or IPO. Ongoing operations and funding rounds are not considered a metric for startup success. **Hyping markers** are elements of the narrative where promotional intensity reaches its peak, amplifying ambition, urgency, or superiority. They typically appear as buzzwords, jargon, or acronyms that project confidence by reducing semantic transparency. Their goal is to mitigate information asymmetry, shape external perception, and align with investor expectations. Hyping markers are operationalized through the creation of custom-built dictionaries as measurable features, enabling their systematic identification and quantification across the Statement variable.

### C. Feature Extraction

Feature extraction consisted of computational modeling via the narrative mapping outlined earlier, resulting in 850 features carrying both framing and signaling weight. Except for contextualized embeddings all methods used a min-max or binary scaling to ensure comparability and interpretability across features. A summary of feature extraction via narrative mapping is displayed in TABLE III.

TABLE III.

**Feature Extraction via Narrative Mapping**

| *Narrative* | *Variable* | *Semantics* | *Extracted Feature (Count)* | *Extraction Technique* | *Feature Type* |
|---|---|---|---|---|---|
| Descriptors (text) | Statement | Hyping markers | Acronyms Density (1) | Occurrence + Ratio | Continuous |
| | | | Buzzwords Density (1) | Occurrence + Ratio | Continuous |
| | | | Jargon Words Density (1) | Occurrence + Ratio | Continuous |
| | | | Acronyms (88) | Occurrence + Ratio | Discrete |
| | | | Buzzwords (145) | Occurrence + Ratio | Discrete |
| | | | Jargon Words (127) | Occurrence + Ratio | Discrete |
| | | | Frequent Words Density (1) | Occurrence + Ratio | Continuous |
| | | | Adjective Density (1) | Occurrence + Ratio | Continuous |
| | | | Noun Density (1) | Occurrence + Ratio | Continuous |
| | | | Value Density (1) | Occurrence + Ratio | Continuous |
| | | | Verb Density (1) | Occurrence + Ratio | Continuous |
| | | | Statement Length (1) | Extension | Continuous |
| | | Non-hyping markers | Location Mention (1) | Detection | Binary |
| | | | Website Mention (1) | Detection | Binary |
| | | | Founding Year Mention (1) | Detection | Binary |
| | | | Embeddings (384) | Cosine similarity | Vector Dimension |
| | Name | | Name Length (1) | Extension | Continuous |
| | | | Single Alphabet Letters (26) | Occurrence + Ratio | Continuous |
| | Website URL | | Website Length (1) | Extension | Continuous |
| | | | Website/Name Length (1) | Equivalence | Binary |
| | | | Website with .com (1) | Detection | Binary |
| Context (category) | Industry | | Industry Tags (61) | Detection | Discrete |
| | | | Industry Tags Count (1) | Occurrence | Continuous |
| | Country | | California vs. US vs. Other (1) | Proportion | Discrete |
| | | | Country (1) | Proportion | Discrete |
| | Age | | Age (1) | Extension | Discrete |

### D. Subsets

The final step of this analysis is the identification of twelve distinct data subsets based on the extracted features. The subsets present data at the Strategy, Variable, Semantics and Feature level, and are used in multiple experiments for benchmarking, validation, and comparison against the baseline subset Descriptors.

- **Narrative-level subsets** consist in features from Context and Descriptors, plus an All-Features. The last two are tested with and without Embeddings.
- **Variable-level subsets** consist in features from the Statement, Name, Website URL, Industry, and Country variables, evaluating single narrative components. The Age variable is tested individually at feature-level.
- **Semantics-level subsets** consist in features from the hyping markers (conveying hype from Statement variable), and the non-hyping markers (enabling the evaluation by contrast).
- **Feature-level subsets** consist in features from Age and Embeddings (384 dimensions but considered as a single feature).

## IV. EXPERIMENTS

### A. Distance Correlation

This experiment evaluates the statistical significance of the subsets toward the target variable Exit. A Python algorithm is built on purpose, applying the dcor open-source library to quantify:

- Significance of each subset (p-value and t-stat).
- Suitability of Descriptors subset as target.
- Value of adding Embeddings to text-based compared to context-based subsets.

Subsets correlation to Exit (if p-value<0.05) and their evidence (t-stat) are presented in TABLE IV:

- All subsets have correlation except Name, Age has the highest t-stat.

- Descriptors (incl. Embeddings) supports our hypothesis with mid-range t-stat.
- If Embeddings are added to text-only subsets the t-stat is stronger: Descriptors (+28%), Statement (+55%), and hyping markers (+22%). Conversely, the hybrid text/category subsets show semantic dilution: non-hyping markers (–77%) and All Features (–63%).

TABLE IV.

| Statistical significance of subsets | | | |
|---|---|---|---|
| ***Subset (excl. Embeddings)*** | ***Features*** | ***p-value*** | ***t-stat*** |
| Age | 1 | 0.0000 | 872.5510 |
| Context | 64 | 0.0000 | 150.9553 |
| Non-hyping markers | 97 | 0.0000 | 147.7209 |
| All Features | 466 | 0.0000 | 136.8623 |
| Industry | 62 | 0.0000 | 69.8190 |
| Descriptors | 402 | 0.0000 | 41.2532 |
| Statement | 372 | 0.0000 | 22.7787 |
| Hyping markers | 369 | 0.0000 | 39.6195 |
| Website URL | 3 | 0.0000 | 48.5760 |
| Country | 2 | 0.0000 | 42.6603 |
| Name | 27 | 0.1212 | 1.1691 |
| ***Subset (incl. Embeddings)*** | ***Features*** | ***p-value*** | ***t-stat*** |
| Non-hyping markers | 481 | 0.0000 | 83.5771 |
| All Features | 850 | 0.0000 | 83.9120 |
| Descriptors | 786 | 0.0000 | 56.9074 |
| Statement | 756 | 0.0000 | 50.6187 |
| Hyping markers | 753 | 0.0000 | 50.5496 |
| Embeddings | 384 | 0.0000 | 49.6230 |

## B. Subset Importance

This experiment assesses the relative predictive power scale of each subsets in determining Exit. A Python-based Random Forest Classifier algorithm is built on purpose, applying the scikit-Learn open-source library to quantify importance ranking of narrative-, variable-, semantics-, and feature-level subsets. The subsets are evaluated without Embeddings to avoid inexplicability. Subsets importance is presented in TABLE V:

- At **narrative-level**, Descriptors carries a much larger importance than Context despite the latter includes Age, the feature most correlated with Exit.
- At **variable-** and **semantics-level**, both Statement and hyping markers capture high importance, indicating the value of hype-based textual features when categorical features are not included, and validating the assumption that textual features have predictive power.
- **Individual feature scores** confirm that density-based with hype potential features dominate the ranking, except for Age, which remains the single most important individual feature for predicting Exit.

TABLE V.

| Subsets Importance Analysis | | |
|---|---|---|
| ***Narrative-level Subsets*** | ***Feature Share*** | ***Importance*** |
| Descriptors | 86.3% | 0.7959 |
| Context | 13.7% | 0.2041 |
| ***Variable-level Subsets*** | ***Feature Share*** | ***Importance*** |
| Statement | 79.8% | 0.5484 |
| Name | 5.8% | 0.2214 |
| Age | 0.2% | 0.1104 |
| Industry | 13.3% | 0.0759 |
| Website URL | 0.6% | 0.0261 |
| Country | 0.4% | 0.0178 |
| ***Semantics-level Subsets*** | ***Feature Share*** | ***Importance*** |
| Hyping markers | 79.2% | 0.5454 |
| Non-hyping markers | 20.8% | 0.4546 |
| ***Feature-level Subsets**** | ***Feature Share*** | ***Importance*** |
| Age | 0.2% | 0.1104 |
| Statement Length | 0.2% | 0.0332 |
| Noun Density | 0.2% | 0.0230 |
| Verb Density | 0.2% | 0.0208 |
| Frequent Words Density | 0.2% | 0.0205 |
| Buzzwords Density | 0.2% | 0.0187 |
| Jargon Word Density | 0.2% | 0.0181 |
| Adjective Density | 0.2% | 0.0178 |

* Top 8 importance scores

## C. Exit Ratio Thresholds

This experiment analyzes individual features to assess the Exit Ratio thresholds for each. Features are divided into three types: discrete (e.g., Country), binary (e.g., Location Mention). and continuous (e.g., Buzzword Density). Each type is sampled and evaluated with different methods. We demonstrate that a selective optimization of binary and continuous features can increase the chances of Exit compared to a linear application.

The experiment quantifies:

- Exit Ratio for discrete, binary and continuous features.
- Exit Delta against the global Exit average.
- Maximum Exit Delta for binary and continuous features with selective optimization.

Definitions:

- **Exit Ratio**: proportion of exits in the sample.
- **Exit Delta**: Exit Ratio change compared to the global Exit average (10.38%).

### 1) Discrete Features

For each category, the proportion relative to the total is calculated, and Exit Ratio and Exit Delta are cross-referenced for each category. For Country and Industry Tags, the statistically significant sample size is determined through 95% confidence interval. For the Age feature, categories are grouped into four ranges. Exit Ratio for discrete features are presented in TABLE VI:

- Country feature presents North America (Canada and United States) having the highest Exit Ratio and the only positive Exit Delta.
- The first three Industry Tags categories are related to leisure and B2C startups (Travel, Media & Telecom, and Gaming), rather than B2B or productivity.
- Age categories are consistent with industry standards: only a few Exits in the 1-5 years range, and increasing sharply after 5 years in operations.

TABLE VI.

| Exit Ratio for Discrete Features | | | |
|---|---|---|---|
| ***Country Variable**** | ***Sample Size*** | ***Exit Ratio*** | ***Exit Delta*** |
| United States | 4268 | 12.40% | 19.45% |
| Canada | 293 | 10.40% | 0.18% |
| Germany | 112 | 9.68% | –6.76% |
| Israel | 103 | 8.04% | –22.58% |
| ***Industry Tags Variable**** | ***Sample Size*** | ***Exit Ratio*** | ***Exit Delta*** |
| Travel | 227 | 18.75% | 80.64% |
| Media & Telecom | 114 | 14.09% | 35.74% |
| Gaming | 276 | 9.90% | –4.62% |
| Human Resources | 296 | 9.33% | –10.12% |
| ***Age Variable*** | ***Sample Size*** | ***Exit Ratio*** | ***Exit Delta*** |
| 15-20 years | 61 | 52.5% | 406.2% |
| 10-15 years | 624 | 43.9% | 323.5% |
| 5-10 years | 2189 | 19.5% | 88.2% |
| 1-5 years | 4545 | 2.1% | –80.0% |

* Top 4 Exit Ratio scores

### 2) *Binary Features*

Features with binary normalization produced only two samples, that are cross-reference with the number of Exits, without applying statistical significance testing, since both are relevant. Results are presented in TABLE VII:

- In Statement Variable, mentions of Website and Founding Year shows higher Exit Delta. Mention of Location shows lower Exit Delta.
- Different Website/Name Length and use of a non-.com website produce a lower Exit Delta.

Selective optimization (i.e., the average of best Exit Delta for each feature) can increase the chances of Exit compared to a linear application (i.e., all features 0 or 1).

**Exit Delta - Statement Variable**

- Linear application (all features 0): 0.3%
- Linear application (all features 1): 10.4%
- Optimized application: 19.9%

**Exit Delta - Website URL Variable**

- Linear application (all features 0): –29.6%
- Linear application (all features 1): 16.4%
- Optimized application: 16.4%

Optimized application based on Exit Delta shows higher performance than uniform values (all 0 or all 1). For instance, when all values equal 1 (Location, Founding Year, and Website included in the Statement), the average Exit Delta is 10.4%, but it nearly doubles to 19.9% when Location is absent. This likely reflects cases where startups present themselves as local rather than global solutions, a framing less favorable for Exit. In contrast, adopting a ".com" domain and aligning Name Length with Website Length yield the highest Exit Delta.

TABLE VII.

| Exit Ratio for Binary Features | | | |
|---|---|---|---|
| ***Statement Variable*** | ***Sample Size*** | ***Exit Ratio*** | ***Exit Delta*** |
| Location Mention | | | |
| 0 | 6794 | 10.6% | 2.4% |
| 1 | 625 | 7.7% | –26.0% |
| Founding Year Mention | | | |
| 0 | 7283 | 10.3% | -0.7% |
| 1 | 136 | 14.0% | 34.6% |
| Website Mention | | | |
| 0 | 7152 | 10.3% | –0.9% |
| 1 | 267 | 12.7% | 22.7% |
| ***Website URL Variable*** | ***Sample Size*** | ***Exit Ratio*** | ***Exit Delta*** |
| Website/Name Length | | | |
| 0 | 2309 | 7.7% | –26.1% |
| 1 | 5110 | 11.6% | 11.8% |
| Website with .com | | | |
| 0 | 2877 | 7.0% | –33.0% |
| 1 | 4542 | 12.6% | 20.9% |

### 3) *Continuous Features*

Each feature is normalized to a 0-1 range, with values grouped into samples that are cross-referenced with the number of Exits. The statistically significant sample size is determined through 95% confidence interval. For each feature, the max, min, and best Exit Delta values are identified. Results are presented in TABLE VIII:

- Acronym and Noun Density perform best at minimum level (0-1% of the statement), indicating that static or technical wording reduce chances of Exit.
- Buzzword and Jargon Density perform best at maximum or near-maximum levels (21-36% of the statement), underscoring the impact of hyped language.
- Frequent Word Density reaches the highest Exit Delta (172%) across the cohort when maximized (68-82% of the statement), highlighting the importance of hyped and trendy language.
- Adjective, Verb, and Value Density perform best at maximum levels (17-54% of the statement), suggesting the effectiveness of dynamic and persuasive language.
- No density measures at the minimum or maximum level produce a negative Exit Delta, implying that extreme narrative perform better than moderate. In contrast, length measures show positive Exit Delta at minimum and negative at maximum, particularly Website Length (–84.7%), consistent with the premium placed on brevity in startup and VC communication.

TABLE VIII.

| Exit Ratio for Continuous Features | | | | | | | |
|---|---|---|---|---|---|---|---|
| ***Density Measure*** | ***MIN*** | ***Exit Ratio*** | ***Exit Delta*** | ***MAX*** | ***Exit Ratio*** | ***Exit Delta*** | ***BEST*** |
| Acronym Density | 0-1% | 11.2% | 7.9% | 10-12% | 10.3% | -0.3% | MIN |
| Buzzword Density | 0-3% | 15.3% | 47.5% | 36-45% | 10.5% | 1.4% | 30-36% |
| Jargon Word Density | 0-3% | 10.8% | 3.7% | 21-27% | 19.7% | 89.9% | MAX |
| Frequent Word Density | 0-7% | 18.8% | 80.9% | 68-82% | 28.2% | 172.1% | MAX |
| Adjective Density | 0-4% | 12.5% | 20.1% | 29-36% | 22.8% | 119.6% | MAX |
| Noun Density | 12-22% | 23.4% | 125.8% | 100.0% | 20.8% | 100.2% | MIN |
| Value Density | 0-2% | 10.9% | 4.9% | 13-17% | 17.1% | 64.5% | MAX |
| Verb Density | 0-5% | 19.6% | 88.7% | 45-54% | 23.1% | 122.4% | MAX |
| ***Length Measure*** | ***MIN*** | ***Exit Ratio*** | ***Exit Delta*** | ***MAX*** | ***Exit Ratio*** | ***Exit Delta*** | ***BEST*** |
| Statement Length | 2-37 words | 11.6% | 11.7% | 151-180 words | 8.1% | –22.1% | MIN |
| Name Length | 3-6 chars | 11.1% | 7.2% | 15-19 chars | 7.9% | –23.6% | MIN |
| Website Length | 3-4 chars | 11.4% | 9.8% | 18-20 chars | 1.6% | –84.7% | 16-17 chars |

For density measures, optimized application outperforms both all MIN and all MAX, indicating that while the best outcomes lean toward high density, a fully maximized hyped narrative is suboptimal. For length measures, optimized application similarly exceeds both all MIN and all MAX, showing that balanced brevity outperforms either extremes.

**Exit Delta - Density Measures**

- Linear application (all features MIN): 47.4%
- Linear application (all features MAX): 83.7%
- Optimized application (all features BEST): 94.4%

**Exit Delta - Length Measures**

- Linear application (all features MIN): 9.6%
- Linear application (all features MAX): –43.5%
- Optimized application (all features BEST): 16.6%

### D. *Predictive Algorithms*

Results from distance correlation show that the Descriptors (text-based) and Context (category-based) subsets carry statistically significant power to predict Exit.

- **Context:** benchmark for assessing the predictive power of text-based subsets in its absence.
- **All features:** benchmark for assessing the impact of combining text and category features. It's evaluated with and without Embeddings.
- **Descriptors:** target subset of our hypothesis, carrying the greatest framing weight. It's evaluated with and without Embeddings to assess if their addition amplifies or reduces predictive power.

**This experiment evaluates whether textual descriptors have standalone predictive power for startup success.** It has two steps. First, six machine-learning models are applied to the subsets, with adjustments for class imbalance and optimization toward maximizing the F1-score. This step evaluates the predictive performance of the subsets and models. Logistic Regression achieves the highest F1-score for the All Features subset excluding Embeddings. Second, a two-step Logistic Regression procedure is applied to the subsets to evaluate Exit predictions at a threshold selected to target precision of 1.0 on validation data. This threshold is then applied to the test data. A recall above 0.01 is treated as a meaningful outcome, representing 1% of actual Exits identified.

#### 1) *F1-score and Subsets Evaluation*

A multi-model algorithm is implemented in Python using four machine-learning models from the scikit-learn library: Naive Bayes, Logistic Regression, Neural Network (NN), and Support Vector Machine (SVM). Each model is configured to address class imbalance, with or without SMOTE oversampling. LightGBM and XGBoost are also evaluated.

K-fold cross-validation assesses generalization: the dataset is split into k folds, the model trains on k-1 folds and tests on the remaining fold, and this process repeats until each fold has served as the test set. Results are averaged to reduce dependence on a single partition. For F1-score, the probability threshold is selected on the first fold to maximize F1 and is fixed for the remaining folds. In each run, stratified folds are generated, SMOTE is applied, the model is trained, predictions are made, the threshold converts probabilities to binary labels, and F1, ROC AUC, precision, and recall are calculated and averaged. Results are presented in TABLE IX:

- Descriptors subset shows standalone predictive performance.
- Logistic Regression achieves the best F1-score overall and is the model selected for the Exit prediction experiment.
- Context and All Features subsets have the highest F1-scores, showing that category-based and hybrid text/category subsets have most predictive power.
- Adding Embeddings in hybrid text/category subsets (All-Features), introduces noise, reducing performance. This pattern is consistent with the distance correlation test results. In text-based subsets the addition of Embeddings have marginal benefits.

TABLE IX

| F1-score and Subsets Evaluation | | | | | |
|---|---|---|---|---|---|
| ***Subset*** | ***Model*** | ***Average F1*** | ***Average Precision*** | ***Average Recall*** | ***Average ROC AUC*** |
| | Naive FixedProb | 0.1881 | 0.1038 | 1 | 0.5000 |
| Context | Logistic Regression | 0.4703 | 0.3776 | 0.6247 | 0.8603 |
| t-stat = 150.9553 | SVM | 0.4577 | 0.3992 | 0.5377 | 0.8518 |
| | Neural Network | 0.4031 | 0.3106 | 0.5779 | 0.8153 |
| | XGBoost | 0.4426 | 0.4387 | 0.4494 | 0.8473 |
| | LightGBM | 0.4569 | 0.4118 | 0.5169 | 0.8538 |
| All Features excl. Embeddings | Logistic Regression | 0.4830 | 0.4286 | 0.5545 | 0.8576 |
| t-stat = 136.8623 | SVM | 0.4584 | 0.4095 | 0.5221 | 0.8501 |
| | Neural Network | 0.3911 | 0.3124 | 0.5247 | 0.7758 |
| | XGBoost | 0.4368 | 0.3283 | 0.6532 | 0.849 |
| | LightGBM | 0.4595 | 0.351 | 0.6662 | 0.8606 |
| All Features incl. Embeddings | Logistic Regression | 0.4306 | 0.3646 | 0.5286 | 0.8379 |
| t-stat = 83.9120 | SVM | 0.4146 | 0.4382 | 0.3974 | 0.8315 |
| | Neural Network | 0.3762 | 0.3399 | 0.4234 | 0.7831 |
| | XGBoost | 0.456 | 0.3692 | 0.5987 | 0.8515 |
| | LightGBM | 0.4465 | 0.4124 | 0.4883 | 0.8541 |
| Descriptors incl. Embeddings | Logistic Regression | 0.2623 | 0.1686 | 0.5935 | 0.6789 |
| t-stat = 56.9073 | SVM | 0.2394 | 0.287 | 0.2078 | 0.6828 |
| | Neural Network | 0.2284 | 0.1811 | 0.3104 | 0.6192 |
| | XGBoost | 0.2495 | 0.1969 | 0.3416 | 0.6383 |
| | LightGBM | 0.2593 | 0.1713 | 0.5338 | 0.6534 |
| Descriptors excl. Embeddings | Logistic Regression | 0.2607 | 0.2213 | 0.3182 | 0.6581 |
| t-stat = 41.2532 | SVM | 0.2511 | 0.2178 | 0.2987 | 0.6604 |
| | Neural Network | 0.2193 | 0.1564 | 0.3675 | 0.5902 |
| | XGBoost | 0.2544 | 0.1699 | 0.5078 | 0.6519 |
| | LightGBM | 0.2718 | 0.2167 | 0.3662 | 0.6502 |

*2) Exit Prediction*

A two-step Logistic Regression procedure is implemented in Python using scikit-learn. A class-balanced Logistic Regression model is evaluated using stratified 10-fold cross-validation. Within each fold, the training portion is split into an 80% model-training subset and a 20% validation subset. A probability threshold targeting precision of 1.0 for the positive class (Exit = 1) is selected using the validation subset and then applied to the test fold. Exit = 1 is evaluated by recall and F1; Exit = 0 is evaluated using a fixed 0.5 threshold.

The model is configured with class balancing and up to 1,000 iterations. Logistic Regression is evaluated on the subsets using 10-fold stratified cross-validation to predict the binary target Exit. For each run, the data are divided into 10 stratified folds. One fold is held out as the test subset; the remaining nine folds form the training portion. Within that training portion, 80% is used to fit the model and 20% is used for validation. A probability threshold targeting precision of 1.0 for the positive class (Exit = 1) is selected using the validation subset. The threshold is then applied to the held-out test subset, and recall and F1 are recorded for Exit = 1. Because the threshold is selected on validation data, precision of 1.0 is a target rather than a guarantee for the test subset; test precision is measured from the resulting predictions. For the non-Exit evaluation, the predictions are assessed using a fixed 0.5 threshold to estimate precision, recall, and F1 for the negative class (Exit = 0).

The purpose of this procedure is to examine the recall associated with a validation-selected threshold targeting precision of 1.0 for Exit predictions, while also reporting performance for the non-Exit class. Results are presented in TABLE X:

- **Descriptors** incl. Embeddings subset identifies 1% of Exits (recall = 0.010) with observed precision of 1.0 across 10 of 10 folds, while also capturing non-Exit cases (recall = 0.253).
- **Embeddings** act as a predictive amplifier within text variables, doubling Descriptors recall from 0.005 to 0.010 when introduced, but introduce structural noise in hybrid text/categories variables, cutting All Features recall from 0.017 to 0.008.
- **Robustness** is assessed using 5-, 10-, 15-, and 20-fold configurations. For the Descriptors subset, observed precision is 1.0 in all reported configurations; recall varies across configurations.

TABLE X

| Exit Prediction | | | | | |
|---|---|---|---|---|---|
| ***Subset*** | ***Class*** | ***Precision*** | ***Recall*** | ***F1-score*** | ***Support*** |
| All Features excl. Embeddings | Exit | 1 | 0.017 | 0.032 | 770 |
| | non-Exit | 0.688 | 0.183 | 0.289 | 6649 |
| Context | Exit | 1 | 0.014 | 0.027 | 770 |
| | non-Exit | 0.699 | 0.176 | 0.282 | 6649 |
| Descriptors incl. Embeddings | Exit | 1 | 0.010 | 0.020 | 770 |
| | non-Exit | 0.825 | 0.253 | 0.387 | 6649 |
| All Features incl. Embeddings | Exit | 1 | 0.008 | 0.015 | 770 |
| | non-Exit | 0.694 | 0.167 | 0.268 | 6649 |
| Descriptors excl. Embeddings | Exit | 1 | 0.005 | 0.010 | 770 |
| | non-Exit | 0.842 | 0.317 | 0.460 | 6649 |
| **Robustness Check** | | | | | |
| ***Subset Descriptors*** | ***Class*** | ***Precision*** | ***Recall*** | ***F1-score*** | ***Support*** |
| 5 folds | Exit | 1 | 0.040 | 0.080 | 770 |
| | non-Exit | 0.816 | 0.254 | 0.387 | 6649 |
| 10 folds | Exit | 1 | 0.010 | 0.020 | 770 |
| | non-Exit | 0.825 | 0.253 | 0.387 | 6649 |
| 15 folds | Exit | 1 | 0.016 | 0.030 | 770 |
| | non-Exit | 0.821 | 0.253 | 0.387 | 6649 |
| 20 folds | Exit | 1 | 0.014 | 0.027 | 770 |
| | non-Exit | 0.820 | 0.257 | 0.391 | 6649 |

## V. BUSINESS APPLICATIONS

VC firms face challenges separating genuine startup potential from exaggerated claims, as founders often use persuasive language that can obscure true business value. Quantitative hype evaluation aids investors in prioritizing credible startups and improving decision efficiency during screening and due diligence. Measuring hype supports risk mitigation by identifying overstated claims and biases, enabling more rigorous and data-driven investment decisions. Incorporating such metrics enhances startup evaluation accuracy and reduces information asymmetry (CUMMING 2008; CHEMMAUR 2021).

This research enables the development of a Hyping Score as a potential business application, providing a standardized metric to assess hype intensity within startup communications. The Hyping Score quantifies the degree to which startup Descriptors are hyped by founders, reflecting the use of exaggerated persuasion or promotional language that may compromise clarity and accuracy. Specifically, the score incorporates metrics such as statement length; densities of acronyms, buzzwords, jargon, and frequent words; densities of parts of speech including adjectives, nouns, values, and verbs; and the presence and ratio of acronyms, buzzwords, and jargon identified through customizable dictionaries. The dictionaries used for detecting these elements can be tailored to capture the specific language preferences, nuances, or hype thresholds relevant to a particular VC firm, industry sector, or hype cycle stage. This customization allows the Hyping Score to serve as a benchmarking tool by comparing startup narratives against selected reference ranges. Hyping Score is calculated as the average of normalized hyping markers values multiplied by their relative importance:

$$\text{Hyping Score} = \frac{1}{N}\sum_{i=1}^{N} h_i\,(s_i)$$

Where:

- $N$ = number of hyping markers
- $h$ = normalized values of hyping markers
- $s$ = hyping markers importance

## VI. CONCLUSIONS

Early-stage founders use framing and signaling to persuade VC firms who must evaluate opportunities under severe information asymmetry and large application volume. In this context, computational prediction of startup success is valuable, and this research tests whether narratives alone carry predictive power. Unlike prior work, it focuses on text-only features within VC-validated datasets and applies a stricter definition of success – cash-returning Exit – rather than proxies such as funding or continued operations, making results harder to obtain but more relevant to investors and founders.

This study concludes that text-based Descriptors can independently predict startup success without reliance on context, financials, or human capital variables. Exit prediction attained 0.48 F1/0.55 recall using both text-based and category-based features (excl. Embeddings), and 0.26 F1/0.59 recall with Descriptors only (incl. Embeddings). In addition, a further optimized 2-step Logistic Regression using Descriptors only (incl.

Embeddings), attained precision 1 and recall 0.01, indicating 1% correct positive prediction among 770 actual Exit. We have also analyzed vector Embeddings during distance correlation and machine learning prediction, finding their value only in text-only datasets. When combined with other categorical variables, Embeddings reduce predictive power and introduce noise when combined with categorical variables.

A framework for Exit Ratio thresholds and textual hyping markers evaluation was established by engineering features via narrative mapping. Selective optimization of continuous and binary textual features increases success probability compared to linear application. A Hyping Score was outlined, indicating potential for business applications.

Limitations include: (1) specificity to only two US-based VC firms, which may affect generalizability; (2) dictionary-based hyping markers , which may miss context-specific or emerging hype language; and (3) predictive recall intentionally low to guarantee perfect precision.

Our research shows that startup textual descriptors, particularly statements, can predict cash-generating outcomes and be optimized for hyping levels, providing a valuable framework for venture capitalists and founders. The results suggest utility across multiple domains. One application is a SaaS tool to benchmark hype signals, improving efficiency and accuracy in venture capital pitching and due diligence by quantifying narrative strength. Another is supporting RAG systems with domain-specific feature engineering for venture contexts, enhancing precision and relevance of decision-making. Finally, the findings also provide a framework for synthesizing data that mirror startup-specific semantic patterns, enabling probabilistic generation of realistic synthetic datasets to aid open-source analysis, model training, and validation when real data are limited or sensitive.